\def\year{2019}\relax
\documentclass[letterpaper]{article}
\usepackage{aaai}
\usepackage{multirow}
\usepackage{booktabs}
\usepackage{times}
\usepackage{helvet}
\usepackage{courier}
\usepackage{kotex}
\usepackage{graphicx}
\usepackage{color}
\usepackage{amsmath,amssymb}
\usepackage{graphicx}
\usepackage{caption}
\usepackage{subcaption}
\usepackage{breqn}
\usepackage{algorithm}
\usepackage[noend]{algpseudocode}
\DeclareGraphicsExtensions{.pdf,.png,.jpg}
\newcommand{\veryshortarrow}[1][3pt]{\mathrel{%
   \hbox{\rule[\dimexpr\fontdimen22\textfont2-.2pt\relax]{#1}{.4pt}}%
   \mkern-4mu\hbox{\usefont{U}{lasy}{m}{n}\symbol{41}}}}

\newcommand\sm[1]{\textcolor{black}{#1}}
\newcommand\nj[1]{\textcolor{black}{#1}}
\newcommand\hy[1]{\textcolor{black}{#1}}
\newcommand\gj[1]{\textcolor{black}{#1}}

\begin{document}

%
\title{Towards Governing Agent's Efficacy: \\
Action-Conditional $\beta$-VAE for Deep Transparent Reinforcement Learning}

\author{John Yang$^{1}$, Gyujeong Lee$^{1}$, Minsung Hyun$^{1}$, Simyung Chang$^{1,2}$, Nojun Kwak$^{1}$\\
$^{1}$Seoul National University, Seoul, South Korea\\
$^{2}$Samsung Electronics, Suwon, South Korea\\
\{yjohn, regulation.lee, minsung.hyun, timelighter, nojunk\}@snu.ac.kr
}
\maketitle

\begin{abstract}
\begin{quote}
We tackle the blackbox issue of deep neural networks in the settings of reinforcement learning (RL) where neural agents learn towards maximizing reward gains in an uncontrollable way. 
Such learning approach is risky when the interacting environment includes an expanse of state space because it is then almost impossible to foresee all unwanted outcomes and penalize them with negative rewards beforehand. Unlike reverse analysis of learned neural features from previous works, our proposed method \nj{tackles the blackbox issue by encouraging} an RL policy network to learn interpretable latent features through an implementation of a disentangled representation learning method. Toward this end, our method allows an RL agent to understand self-efficacy by distinguishing its influences from uncontrollable environmental factors, which closely resembles the way humans understand their scenes. Our experimental results show that the learned latent factors not only are interpretable, but also enable modeling the distribution of entire visited state space with a specific action condition. We have experimented that this characteristic of the proposed structure can lead to ex post facto governance for desired behaviors of RL agents.

  
 \end{quote}
\end{abstract}

\section{Introduction}


\nj{Despite} many recent successful achievements that \nj{deep neural networks} (DNN) have allowed in machine learning fields \cite{krizhevsky2012imagenet,lecun2015deep,dqn_davidsilver},
the legibility of their high-level representations are noticeably less studied 
compared to the relevant studies which rather prioritize performance enhancements or task completions.
The blackbox issue of neural networks \nj{has been} many times neglected and such technical opacity \nj{has been} excused \nj{for} their vast performance improvements \cite{burrell2016}. 

While the opaqueness of DNN comes handy when strict labels are available for every data sample,
its blackbox issue is a great element of risk especially in reinforcement learning (RL) settings where machines, or agents, are allowed to have highly intertwined interactions with their environments. 
Since an RL agent's policy on action selection is optimized towards maximizing the rewards, 
it may produce harmful and unexpected outcomes if \nj{these} outcomes are not primarily penalized with negative reward signals.

Yet, too much regulation would, contrarily, result in misusing the full potential of the technology \cite{rahwan2018society}.
\nj{RL} is proven of its powerfulness over humans by, for an example of AlphaGo, figuring to learn unprecedented winning moves \cite{silver2017mastering}.
Interfering in the learning process to control the model's resultant \nj{behaviors} 
as done in the work of \cite{humanpreference}
may not be efficient \nj{in governing} RL agents.
\nj{Rather, it is desired to control the efficacy of an agent which is already optimized for the environment.}





In order to rule AI agents efficiently,
humans who govern first need to comprehend how AI machines perceive their world and monitor their efficacy \cite{ml_social_governance,wynne1988unruly}.
\nj{Higgins et al. modeled} an environment with the $\beta$-Variational Autoencoder ($\beta$-VAE) to generate disentangled latent features \nj{\cite{higgins2017darla}}, purposefully inducing the learned features to be interpretable to human \cite{betavae}, and have applied the features for transfer learning across multiple environments. 
We are motivated that \nj{this method}
can be utilized to train an explainable RL agent \cite{higgins2016early}.

We believe building transparent RL agents and governing them would solve issues mentioned above.
In this paper, we propose a method that allows training a deep but transparent RL policy network, encouraging their latent features to be interpretable.
We intend to accomplish this by training RL agents to learn disentangled representations of their world in egocentric perspective with action-conditional $\beta$-VAE (AC-$\beta$-VAE):
the learned \nj{control-dependent} latent features 
and \nj{uncontrollable} environmental factors are disentangled while \nj{the learned factors} are also able to model the environment.
Our strategic design that engage the AC-$\beta$-VAE and \nj{an} RL policy network to share \nj{a} backbone structure overcomes the blackbox issue, supporting the transparency of deep RL.
We also empirically show that the behavior of our agents can further be governed with human enforcements.

\begin{figure*}[ht]
\centering
\begin{subfigure}{.5\textwidth}
  \centering
  \includegraphics[width=1\linewidth]{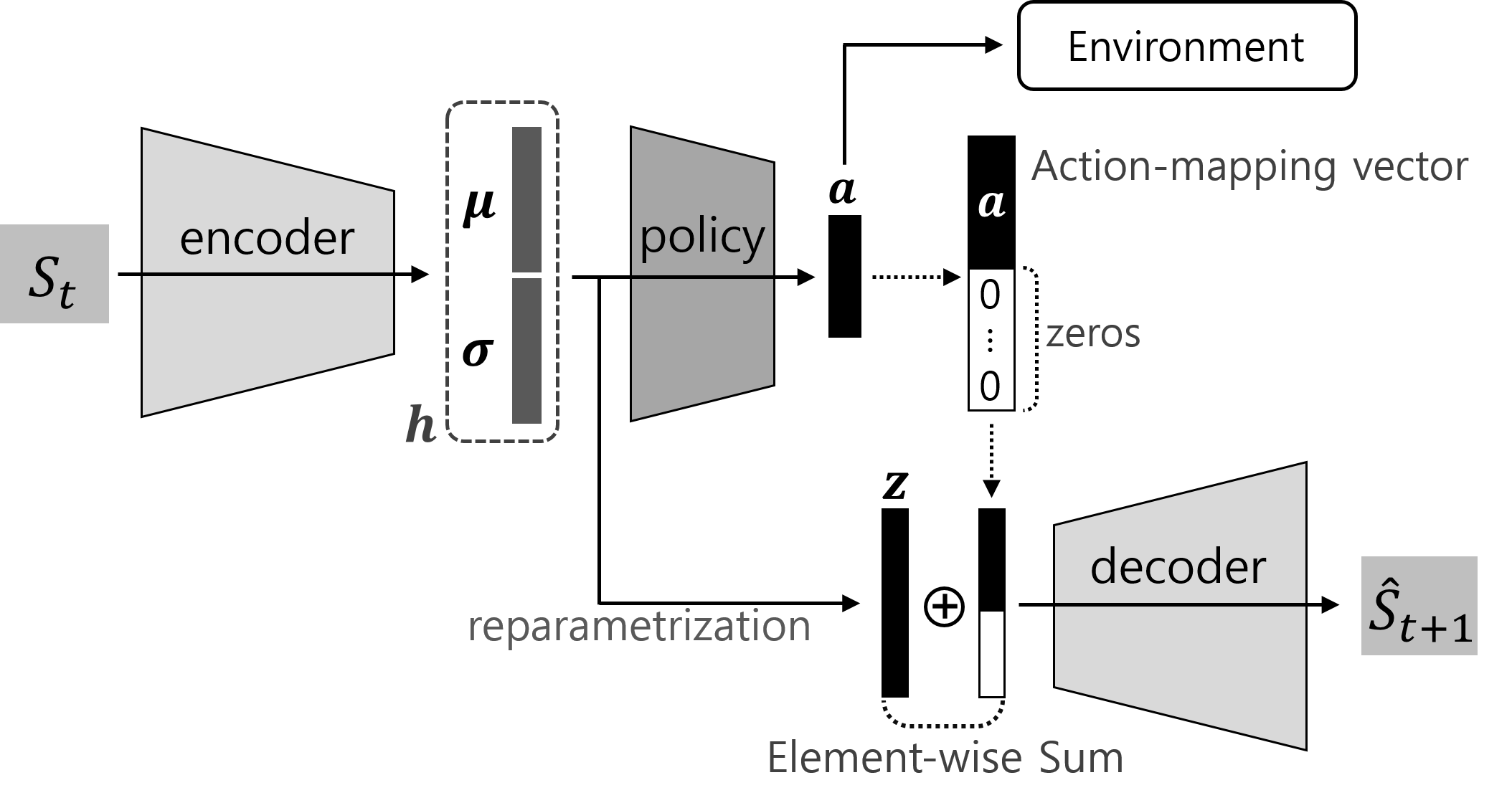}
   \caption{feed flow diagram}
  \label{fig:structure_fflow}
\end{subfigure}%
\begin{subfigure}{.4\textwidth}
  \centering
  \includegraphics[width=1\linewidth]{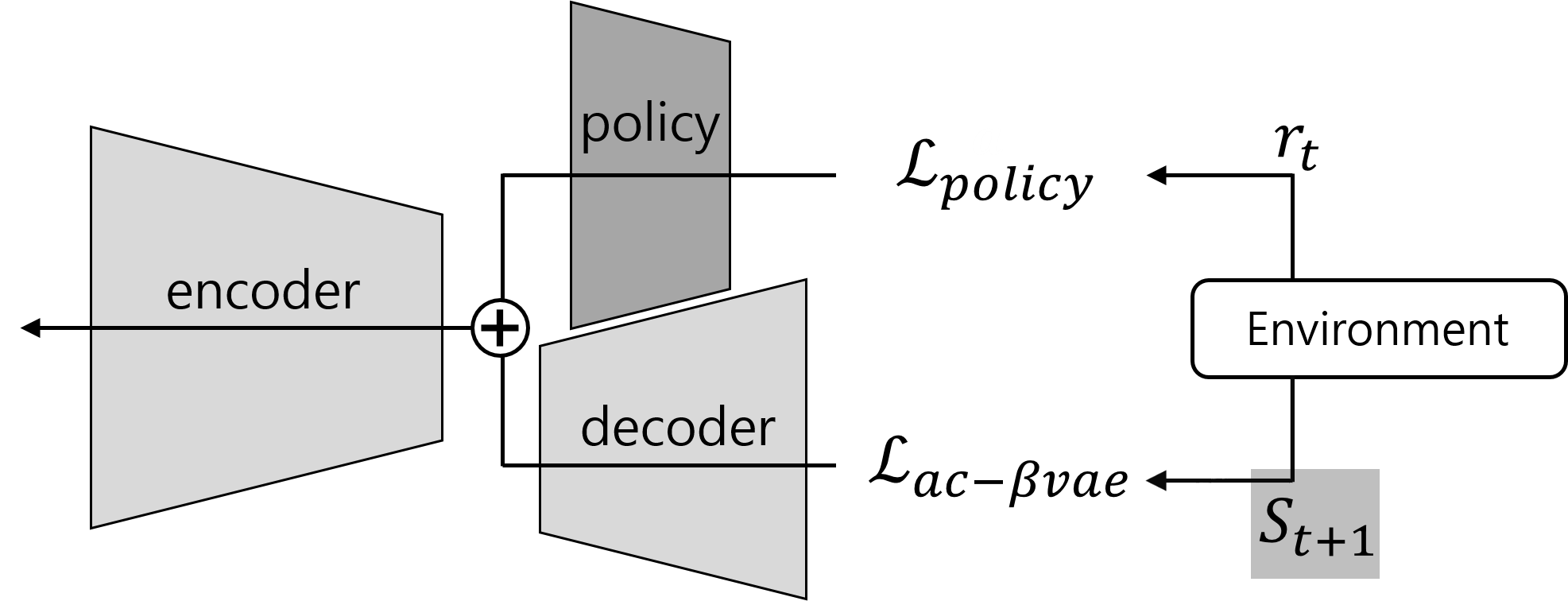}
   \caption{backward flow diagram}
  \label{fig:structure_bflow}
\end{subfigure}
\caption{The structure and flow diagrams of the proposed AC-$\beta$-VAE for a transparent policy network. The proposed network requires training samples of MDP tuples of RL environments that consist of $(s_t, a_t, r_t, s_{t+1})$ where $s_t$, $a_t$ and $r_t$ are respectively state, action and reward at time step $t$. The action-conditional decoder encourages the input features of the policy network to be disentangled and interpretable. Since the encoder + policy network can be seen as one \nj{big} policy network that takes raw states as inputs, its inner intentions in selecting actions \nj{for a desired next state} can thus be explained visually through the outputs of the decoder.
}
\label{fig:structure}
\end{figure*}

\section{Related Work}

Deep learning methods are praised of their unruled pattern extraction that yields better performance in many tasks than machines trained under human prior knowledge
\cite{gunelgooglenet,moore2011autonomous,vanderbilt2012}, but as \nj{stated} earlier, the blackbox characteristic of DNNs can be precarious especially in the RL setting.
One of the safety factors of AI development suggested in \cite{AIsafety} is avoidance of negative side effects when training an agent to complete a goal task with a strict reward function. 

Attempts to open the blackbox of DNN \nj{and} to understand the inner system of neural networks \nj{have been made} in many recent works \cite{lipson2016driverless,zeiler2014visualizing,bojarski2017explaining,visualizing_atari}.
Its inherent learning phenomena are reversely analyzed by observing the \nj{resultant} learned understructure.
While the training progress is also analytically interpreted via information theory \cite{openingbb,openingbb_next},
it is still challenging to anticipate how and why high-level features in neural models are learned in a certain way before training them.
Since learning a disentangled representation
encourages its interpretability \cite{bengio2013representation,betavae}, 
it is previously reported that features of convolutional neural networks (CNN) can also be learned in a visually explainable way \cite{zhang2018visual} through disentangled representation learning.

Prospection of future states conditioned by current actions is meaningful to RL agents in many ways, and action-conditional (variational) autoencoders are learned to predict sequent states in the works of
\cite{worldmodels,oh2015action,thomas2017independently}.
DARLA \cite{higgins2017darla} utilizes disentangled latent representations for cross-domain zero-shot adaptations. \nj{It} aims to prove its representation power in multiple \nj{similar but different} environments. 
Our model may also look similar to conditional generative models like Conditional Variational Autoencoders (CVAE) \cite{cvae} and InfoGan \cite{chen2016infogan}, but these are not directly applicable models to \nj{RL} domains.


\section{Preliminary: $\beta$-VAE}
\nj{Variational autoencoder (VAE) \cite{vae} works} as a generative model based on the distribution of training samples \cite{co2018self,babaeizadeh2017stochastic}.
VAE's goal is to learn the marginal likelihood of a \nj{sample} ${x}$ from a distribution parametrized by generative factors ${z}$. 
In doing so, a tractable proxy distribution $q_\phi({z}|{x})$ 
is used to estimate an intractable posterior $p_\theta({z}|{x})$ with two different parameter vectors $\phi$ and $\theta$.
The marginal likelihood \nj{of a data point ${x}$} can be defined as:
\begin{equation}
\log p_{\theta} ({x}) = D_{KL}(q_\phi ({z}|{x})||p_\theta ({z}|{x})) + L(\theta, \phi, {x}, {z}).
\end{equation}
Since the KL divergence term $D_{KL}(\cdot||\cdot)$ is non-negative, 
\nj{$L_{vae} \triangleq L(\theta, \phi, \bf{x}, \bf{z})$ sets a variational lower bound for the likelihood $\log p_\theta ({x})$ and  
the best approximation $q_\phi ({z}|{x})$ for $p_\theta ({z}|{x})$ can be obtained} 
by maximizing \nj{this lower bound: }
\begin{dmath}
L_{vae} = \mathbb{E}_{q_\phi({z}|{x})}[\log p_\theta({x}|{z})] - D_{KL} (q_\phi({z}|{x})||p({z})).
\end{dmath}
In practice, 
$q_\phi$ and $p_\theta$ are respectively encoder and decoder that are parameterized by deep neural networks, and the prior $p({z})$ is usually set to follow Gaussian distribution \nj{$\mathcal{N}(0,I)$}. The gradients of the lower bound can be approximated using the \textit{reparametrization trick}.

\nj{$\beta$-VAE} \cite{betavae} extends the work and drives VAE to learn disentangled latent features, weighting the KL-divergence term from the VAE loss function \nj{(negative of the lower bound)} with a hyper-parameter $\beta > 1$ :
\begin{equation}
L_{{\beta}vae} = \mathbb{E}_{q_\phi({z}|{x})}[\log p_\theta({x}|{z})] - \beta D_{KL} (q_\phi({z}|{x})||p({z})).
\end{equation}

When $\beta$ is ideally selected and does not severely interfere the reconstruction optimization, 
each latent factor of ${z}$ is learned to be not only independent of each other, but also interpretable.
This means the resultant features follow physio-visual characteristics of our world 
and differ from conventional DNN features that are not so human-friendly.

\section{The Proposed Model}
Our proposed model is composed of two structures: 
a policy gradient RL method and the action-conditional $\beta$-VAE (AC-$\beta$-VAE). 
As shown in Figure \ref{fig:structure}, 
both components are designed to strategically share first layers of \nj{the} encoding network so that the latent features of AC-$\beta$-VAE can also become the input of the policy network. 
\nj{This} simple 
\nj{shared} architecture enables human-level interpretations on behaviors of deep RL methods.

Consider a \nj{reinforcement learning} setting where an actor plays a role of learning policy $\pi_\psi(a_t|s_t)$ and selects an action $a \in \mathcal{A}$ given a state $s \in \mathcal{S}$ at time $t$, and there exists a critic that estimates value of the states $V_w(s)$ to lead the actor to learn the optimal policy. 
Here, $\psi$ and $w$ respectively denote the network parameters of the actor and the critic. 
Training progresses towards the direction of maximizing the objective function based on cumulative rewards, $J(\theta)=\mathbb{E}_{\pi_{\psi}}[\sum_t \gamma^t r_t]$ where $r_t$ is the instantaneous reward at time $t$ and $\gamma$ is a discount factor. 
The policy update objective function \nj{to maximize} is defined as follows:
\begin{equation}
L_{policy} =\mathbb{E}_{\pi}[\log\pi_\psi(s_t,a_t)A^\pi(s_t,a_t)].
\label{eq:pg}
\end{equation}
%
\nj{Here,} 
$A^\pi(s,a)$ is an advantage function, which is defined as it is in asynchronous advantage actor-critic method (A3C) \cite{mnih2016asynchronous}:
\begin{equation*}
A^\pi(s_t,a_t) = 
\sum\limits_{i=0}^{k-1}\gamma^i r(s_{t+i},a_{t+i})+\gamma^k V_w^\pi(s_{t+k})-V_w^\pi(s_t),
\label{eq:advantage}
\end{equation*}
where $k$ denotes the number of steps.
We have used the update method of \textit{Advantage Actor Critic} (A2C) \cite{wu2017a2c}, a synchronous and batched version of A3C,
for Atari domain environments \cite{bellemare2013arcade}.
\textit{Proximal Policy Optimization} (PPO) \cite{Schulman2017PPO} is also used for our experiments in continuous control environments, which reformulates the update criterion with the use of clipping objective constraint $\mathcal{C}$ in the form of:

\begin{equation}
\small
L_{policy} = \mathbb{E}_\pi \left[ \frac{\pi_\psi(a|s)}{\pi_{\psi}^{old}(a|s)} A(s,a) \right] -\mathcal{C} D_{KL}(\pi_{\psi}^{old}(\cdot|s) || \pi_{\psi}(\cdot|s)).
\label{eq:ppo_loss}
\end{equation}
\nj{Here, the subscript $t$ for $a$, $s$ and $A$ is omitted for brevity.}

\subsection{Action-Conditional $\beta$-VAE \nj{(AC-$\beta$-VAE)}}
\nj{As shown in Fig. \ref{fig:structure} } with a given environment, the \nj{policy network combined with the encoder} produces rollouts of typical Markov tuples that
consist of $(s_t, a_t, r_t, s_{t+1})$.
A raw state \nj{$s_t$} feeds into the encoder model and gets encoded into a representation $h \in \mathbb{R}^{2n}$, \nj{where $n$ is the dimension of the the latent space}. 
Since \nj{the policy network and AC-$\beta$-VAE} share the parameters until this encoding process, 
the representation \hy{$h=[\mu^T, \sigma^T]^T$} represents a DNN feature \nj{which is inputted} \nj{to} the policy network while also representing a concatenated form of the mean and the standard deviation vectors $\mu, \sigma \in \mathbb{R}^n$. 
The vectors are reparametrized into a posterior variable $z \in \mathbb{R}^n$ through the AC-$\beta$-VAE pipeline.
The output of the encoder feed-flows into the policy network $\pi(a|h)$ to output an optimal action $a \in \mathbb{R}^m$ where $m<n$ so that an RL environment responds accordingly. 
The action vector \nj{$a$} is then concatenated with a vector of zeros in length of $\mathbb{R}^{n-m}$ to create, we call, an \textit{action-mapping vector} \nj{$a^{map} = [a^T,{0}^T]^T \in \mathbb{R}^n$}.
An element-wise sum of the latent variable $z$ and the action-mapping vector $a^{map}$ is performed in order to map action-controllable factors into the latent vector.
This causes the latent variable sampled to be constrained by the \nj{probability of actions}.
The resultant vector \nj{$z_t + a_t^{map}$} is fed into the decoder network to predict the next state $\hat{s}_{t+1}$.
The prediction is then compared with the real state $s_{t+1}$ given by the environment after the action taken. 
For an MDP tuple collected at time $t$, the loss of AC-$\beta$-VAE is computed 
with the following loss function:

\begin{equation}
\small
\begin{split}
L_{ac-{\beta}vae} = & \mathbb{E}_{q_\phi(h_t|s_t)}[\log p_\theta(s_{t+1}|z_t+a_t^{map})]_{z_t{\sim}\mathcal{N}(h_t)} \\
&- \beta D_{KL} (q_\phi(z_t|s_t)||\mathcal{N}(0,I)).
\end{split}
\label{eq:ac_bvae}
\end{equation}

\begin{algorithm}[tb]
\begin{algorithmic}
\State Initialize encoder $q_\phi(h|s)$ and decoder $p_\theta(s|z)_{z{\sim}\mathcal{N}(h)}$
\State Initialize critic $V_w(s)$, actor $\pi_{\psi}(a|h)$ \nj{and state $s$.}
\While {not stop-criterion}
 \State $t_{start}=t$
 \Repeat
 \State \sm{Take an action $a_t$ with policy $\pi_{\psi}$}
 \State Receive new state $s_{t+1}$ and reward $r_t$
 \Until{ $t-t_{start} \geq \textit{number of step}$} \textbf{or} terminal $s_t$
\State $R =
    \left\{
    \begin{array}{l l}
      0  \quad & \text{for terminal } s_t\\
        V_w(s_t) \quad & \text{for non-terminal } s_t
    \end{array}\right.$
   \For {\sm{ $i \in$ $\{t-1, ..., t_{start}\}$}}
   \State $R \Leftarrow r_i + \gamma R$
   \State Compute $A(s_i, a_i)$ (for A2C or PPO)
   \State Sample $z_i\sim\mathcal{N}(h_i)$ and create $a_i^{map}$
   \State Predict $p_\theta(\hat{s}_{i+1}|z_i + a_i^{map})$
   \State Compute $L_{policy}$ and $L_{ac-{\beta}vae}$  
   \State Update encoder, actor and decoder based on:
   \State \indent $L_{total} = L_{policy} + \alpha L_{ac-{\beta}vae}$ 
   \State Update critic by minimizing the loss:
   \State \indent \nj{$L_{critic}(w)= ( R - V_w(s_i)  )^2$}
  \EndFor

\EndWhile
\end{algorithmic}
\caption{\nj{AC-$\beta$-VAE} with an actor-critic policy network}
\label{alg:ac-bvae-pseudo-code}
\end{algorithm}

As \nj{one can see}, the AC-$\beta$-VAE model can be trained either simultaneously with the policy network or separately, 
and all our experiments are performed with the former because \nj{it is more practical.} 
At each iteration of update, the total \nj{objective function} value is calculated with the weighted sum of \nj{objective function} values from both models:
\begin{equation}
L_{total} = L_{policy} + \alpha L_{ac-{\beta}vae}
\end{equation}
where $\alpha$ is the weight balance parameter. 
Since exploration based on the error between generated outputs and the ground-truths have already been proven on the training enhancement in many RL related works \cite{oh2015action,worldmodels,tang2017exploration}, 
our model rather focuses on feasible training of a transparent neural policy network 
and modeling self-efficacy of agents, not on RL performance improvement.
We thus choose relatively small-valued $\alpha$ \nj{not to} confuse the policy network too much.
A basic pseudo-code for the training scenario of our \nj{proposed} structure is provided in Algorithm \ref{alg:ac-bvae-pseudo-code}.


\begin{figure*}[ht]
\centering
  \includegraphics[width=0.85\linewidth, trim=0cm 0cm 0cm 0.2cm, clip=True]{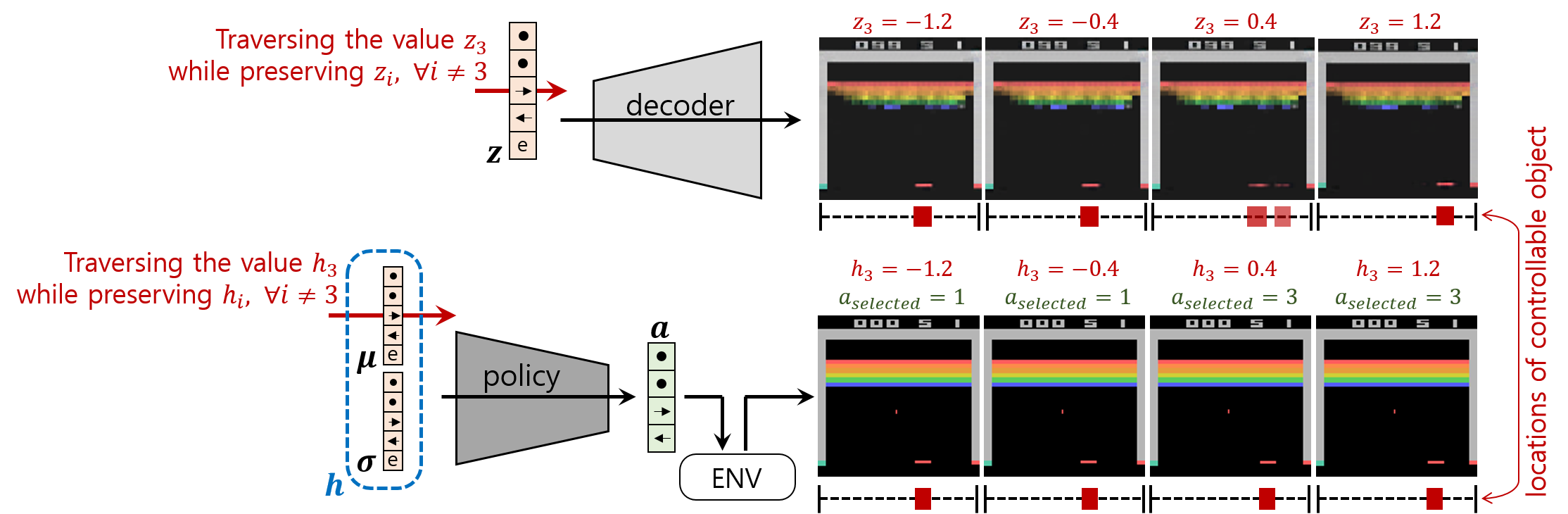}
\caption{The results of traversing \nj{the} latent factor of \nj{our trained} model \nj{on Atari game environment \textsc{Breakout}} with $z \in \mathbb{R}^5$, where $z_{1:4}$ are mapped with variant features of $a \in \mathbb{R}^4$ and $z_{5}$ is condensed with other environmental factors. Since the factors in the latent vector $z$ of AC-$\beta$-VAE are defined by the vectors of mean and standard deviation $\mu, \sigma$, traversing $i$-th value of the latent vector $z_i$ is almost equivalent to traversing $\mu_i$.  The input DNN feature $h$ of the policy network is the concatenation of \nj{$\mu$ and $\sigma$}, and thus \nj{the next state due to} its output actions \hy{$a_{selected}$} caused by traversed $\mu_i$ factor would \nj{be probabilistically predictable by the visual consequence estimated by the decoder with traversed $z_i$}. 
}
\label{fig:transparent_policy}
\end{figure*}

\subsection{Mapping Action-Controllable Representations}
Learning visual influence was previously introduced of its importance and implicitly solved in the works of \cite{oh2015action,visualizing_atari}. 
Distinguishing directly-controllable objects and environment-dependent objects \nj{reflects} much of how a human perceives the world. 
Restricting in the world of Atari game domains as an example, 
it is intuitive for a human agent to first figure out `where I am in the screen' or `what I am capable of with my actions' and then work their ways towards achieving the highest score. 

We show in the \nj{experiment} section that AC-$\beta$-VAE allows RL agents not only to explicitly learn visual influences of their actions, but also learn them in \nj{a} human-friendly way. 
By traversing each \nj{element} of the latent \nj{vector}, we are able to interpret which dimensions are mapped with actions and which are mapped with other environmental factors. 



\begin{figure*}[ht]
\begin{subfigure}{.245\textwidth}
  \captionsetup{justification=centering}
  \includegraphics[width=1.0\linewidth]{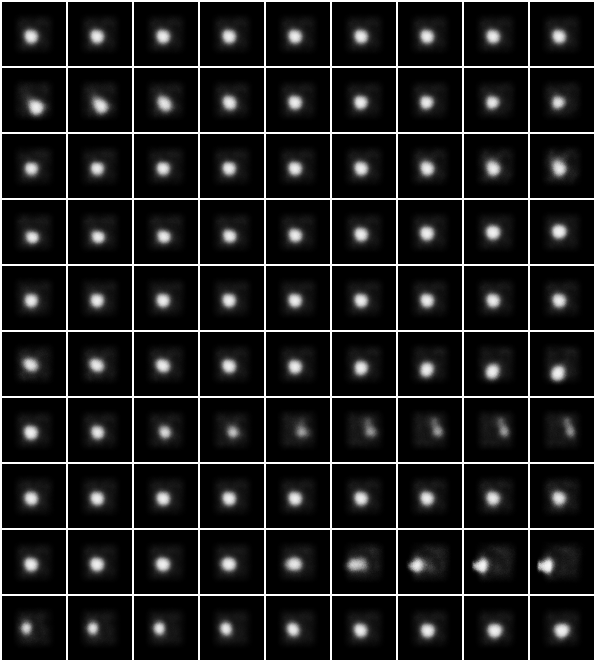}
   \caption{$\beta$-VAE, $\beta$=1 (VAE) \\ without any supervised \\action-mapping}
  \label{fig:dsprites_acbvae_dm}
\end{subfigure}
\begin{subfigure}{.245\textwidth}
  \captionsetup{justification=centering}
  \includegraphics[width=1.0\linewidth]{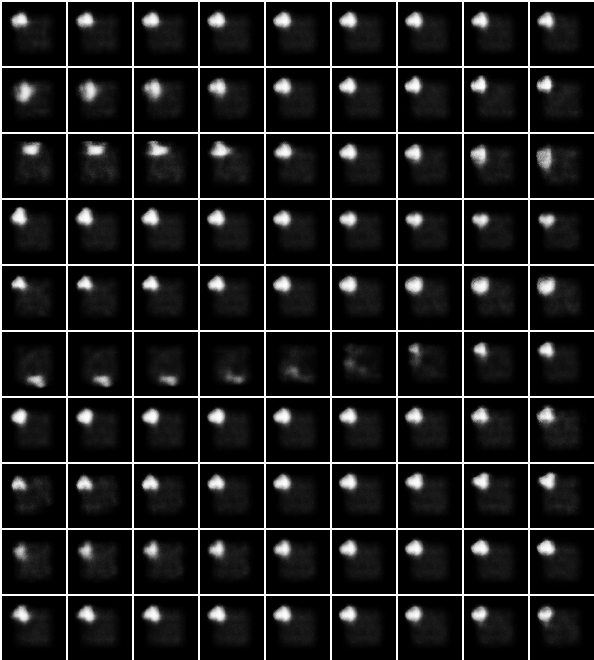}
   \caption{$\beta$-VAE, $\beta$=20 \\ without any supervised \\action-mapping}
  \label{fig:dsprites_bvae}
\end{subfigure}
\begin{subfigure}{.245\textwidth}
  \captionsetup{justification=centering}
  \includegraphics[width=1.0\linewidth]{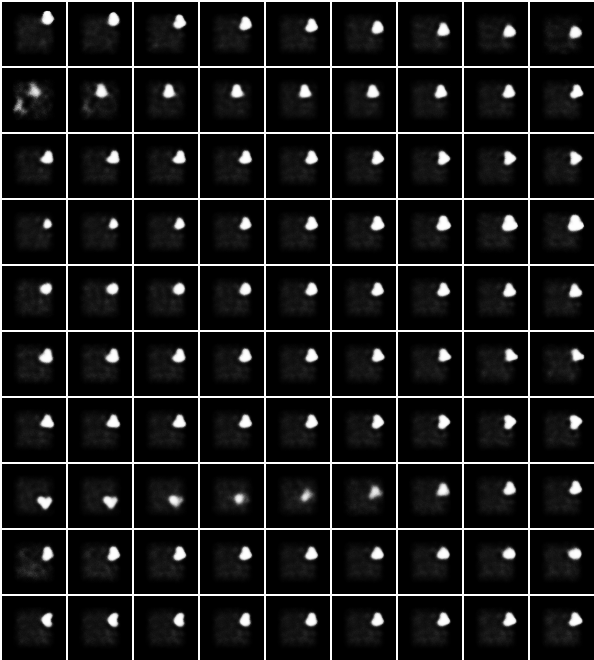}
   \caption{AC-$\beta$-VAE, $\beta$=1 (AC-VAE) with action-mappings :\\
$a_1\veryshortarrow{z_1}$, $a_2\veryshortarrow{z_2}$, $a_3\veryshortarrow{z_3}$, $a_4\veryshortarrow{z_4}$}
  \label{fig:dsprites_acvae}
\end{subfigure}
\begin{subfigure}{.245\textwidth}
  \captionsetup{justification=centering}
  \includegraphics[width=1.0\linewidth]{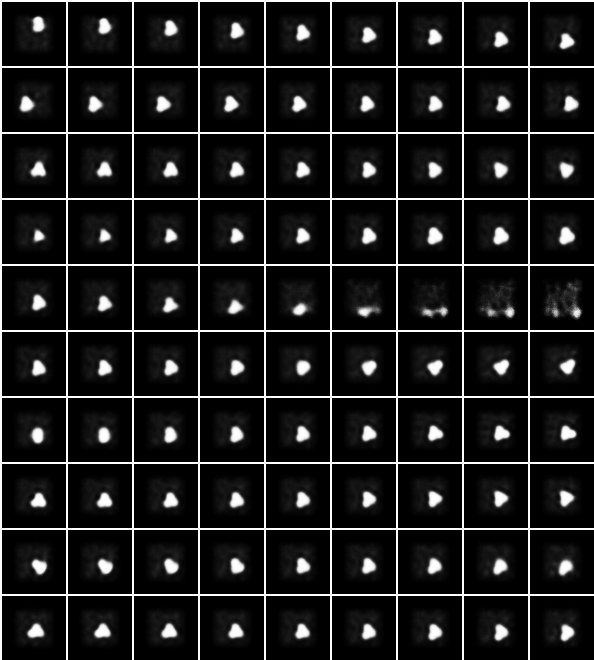}
   \caption{AC-$\beta$-VAE, $\beta$=20\\with action-mappings :\\
$a_1\veryshortarrow{z_1}$, $a_2\veryshortarrow{z_2}$, $a_3\veryshortarrow{z_3}$, $a_4\veryshortarrow{z_4}$}
  \label{fig:dsprites_acbvae}
\end{subfigure}

\caption{The qualitative results of traversing latent factors in $\beta$-VAE with $\beta$=1 (VAE) and \nj{$\beta$=20} on ($s_t, s_{t+1}$) data tuples and those of AC-$\beta$-VAE with $\beta$=1 (AC-VAE) and \nj{$\beta$=20} on $(s_t, a_t, s_{t+1})$ data tuples in dSprites environment. 
The action vectors are retrieved randomly as combinations of ($a_1, a_2, a_3, a_4$) that respectively represent vertical, horizontal, rotational, scaling moves. 
The vertical axes represent the dimensions of the learned latent vector $z_{1:10}$ from top to bottom while the horizontal axes represent traversing values of $[-2:2]$ from left to right.
}
\label{fig:traverse_de}
\end{figure*}

\subsection{Transparent Policy Network}
As mentioned earlier, 
the encoder and the policy network can be grouped as one bigger policy network model with an interpretable layer constrained by the AC-$\beta$-VAE loss.
Unlike high-level features from conventional DNN models, 
the inner features of our policy network are consequentially interpretable.

Figure \ref{fig:transparent_policy} illustrates how our policy network becomes transparent.
If the action-dependent factors are disentangled in the latent vector \nj{$z \in \mathbb{R}^{n}$} and mapped into $z_{1:m}$, 
then so they are in $\mu_{1:m}$ and $\sigma_{1:m}$ because they define the sampling distribution of $z_i$ where $i$ denotes the dimensional location.
The variational samplings from the latent space of VAE is defined as: 
$z_i = \mu_i + \sigma_i \epsilon_i$ 
where $\epsilon$ is an auxiliary noise variable $\epsilon \sim \mathcal{N}(0,1)$. 
And, we know that $q_\phi(z|x)\prod_i dz_i = p(\epsilon)\prod_i d\epsilon_i$.
Since the $\sigma$ value controls mainly the scale of sampled $\epsilon$, traversing $z_i$ is almost equivalent to traversing $\mu_i$\footnote{Refer the original work \nj{of VAE} for more insightful details}. 
Thus, traversing $\mu_i$ encourages the policy network to cause actions as predictions of each traversing value of $z_i$ for $ i \leq m$.

\section{Experiments}
In this section, we present experimental results that demonstrate the following key aspects of our proposed method: 
\begin{itemize}
\item \nj{By mapping actions into the latent vector of $\beta$-VAE, action-controllable factors are disentangled from other environmental factors. 
}

\item \hy{Governance over the optimized behavior of an agent can be made based on human-level interpretation of learned latent behavioral factors. }
\end{itemize}



\noindent We have experimented our method in three different environment types: dSprites, Atari and MuJoCo.

\noindent \textbf{dSprites Environment} is an environment we design with the \textit{dSprites} dataset \cite{dsprites17}.
It originally is a synthetic dataset of 2D shapes that gradually vary in five factors: shape (square, ellipse, heart), scale, orientation, locations in vertical \nj{and horizontal axes, respectively}.
The environment provides a $64 \times 64$ sized image that embrace two shapes, one heart and one square. 
At each time step, the square is randomly scaled in a randomly oriented form at random location within the image. 
The heart-shaped object responds to one of the following discrete action inputs: move upward, downward, left, right, enlarge, shrink, rotate left and right. 
All actions can be represented with a 4-dimensional action vector each of which is responsible for a unit of either vertical, horizontal, scaling or rotating movement.


\begin{table}[bt]
  \centering
  \resizebox{1.\columnwidth}{!}{
    \begin{tabular}{rcccc}
    \toprule
        & {VAE}& {$\beta$-VAE}& {AC-VAE} & {AC-$\beta$-VAE}\\
        &  ($\beta$=1) &  ($\beta$=20)&  ($\beta$=1)&  ($\beta$=20)\\
    
    \midrule
    Avg. Disent. & 0.120 & 0.133 & 0.233 & \textbf{0.390}\\
    Avg. Compl. & 0.155 & 0.231 & 0.288 & \textbf{0.405}\\
    \bottomrule
    \end{tabular}%
  }
    \caption{The quantitative scores of disentanglement and completeness averaged over dimensions of the latent vector learned with $(s_t, a_t, s_{t+1})$ tuples from dSprites environment. 
}
  \label{tab:quant_scores}%
\end{table}%

\begin{figure}[t]
\centering
  \includegraphics[width=.80\linewidth, trim=0.2cm 1.4cm 0cm 0.2cm, clip=True]{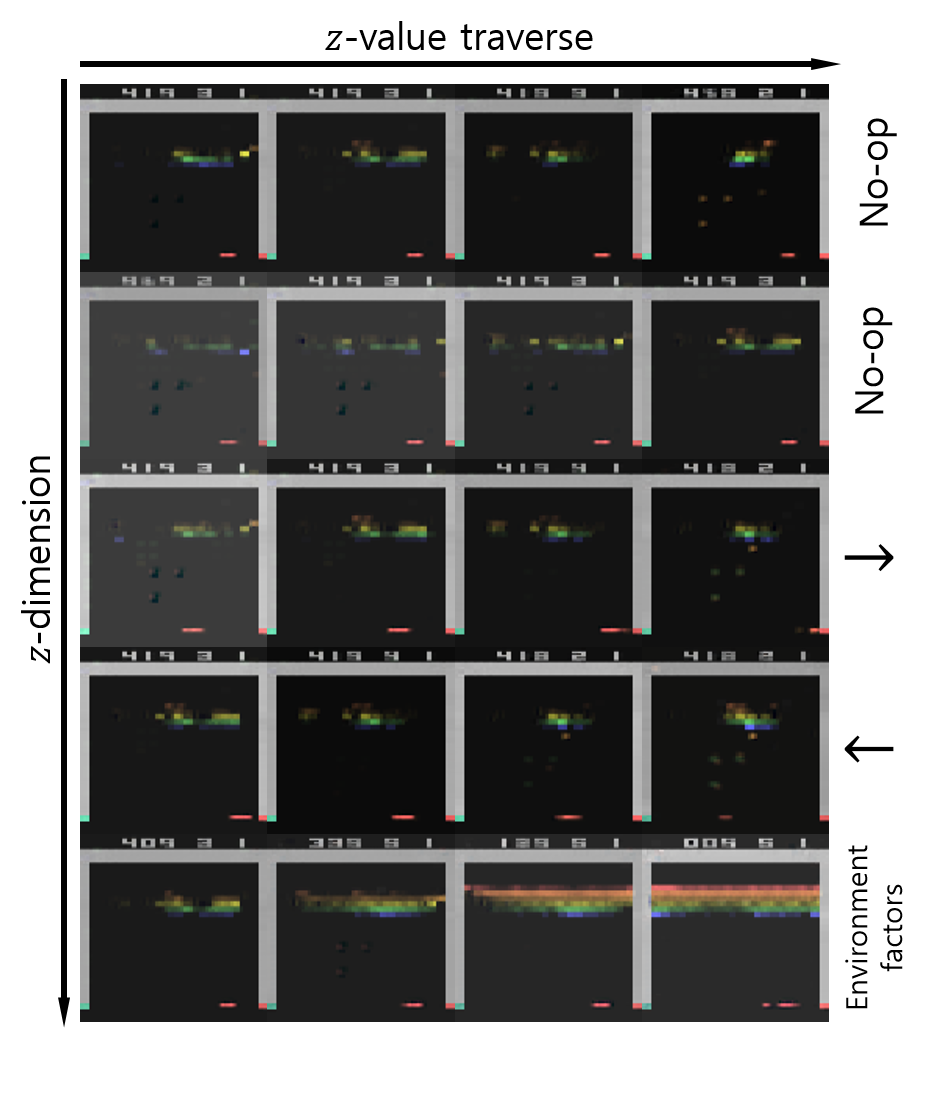}
\caption{The images are \nj{the estimated next states obtained by} traversing the latent vector $z \in \mathbb{R}^5 $ learned by AC-$\beta$-VAE \nj{with $\beta$=10 and $\alpha$=0.001} on the Atari game environment \textsc{Breakout}. \hy{The factors at $z_{1:4}$ are mapped with the control factors such as movements of the paddle, and $z_5$ is mapped with the \nj{environmental} factors \nj{such as} bricks and the scoreboard.}}
\label{fig:ego_worldmodel}
\end{figure}

\noindent \textbf{Atari Learning Environment} is a software framework for assessing RL algorithms \cite{bellemare2013arcade}.
Each frame is considered as a state and immediate rewards are given for every state transitions.
Our method is experimented in the Atari game environments of \textsc{Breakout, Seaquest} and \textsc{Space-Invaders.}

\begin{figure}[t]
\centering
  \includegraphics[width=0.87\linewidth, trim=0.05cm 0.05cm 0.2cm 0.2cm, clip=True]{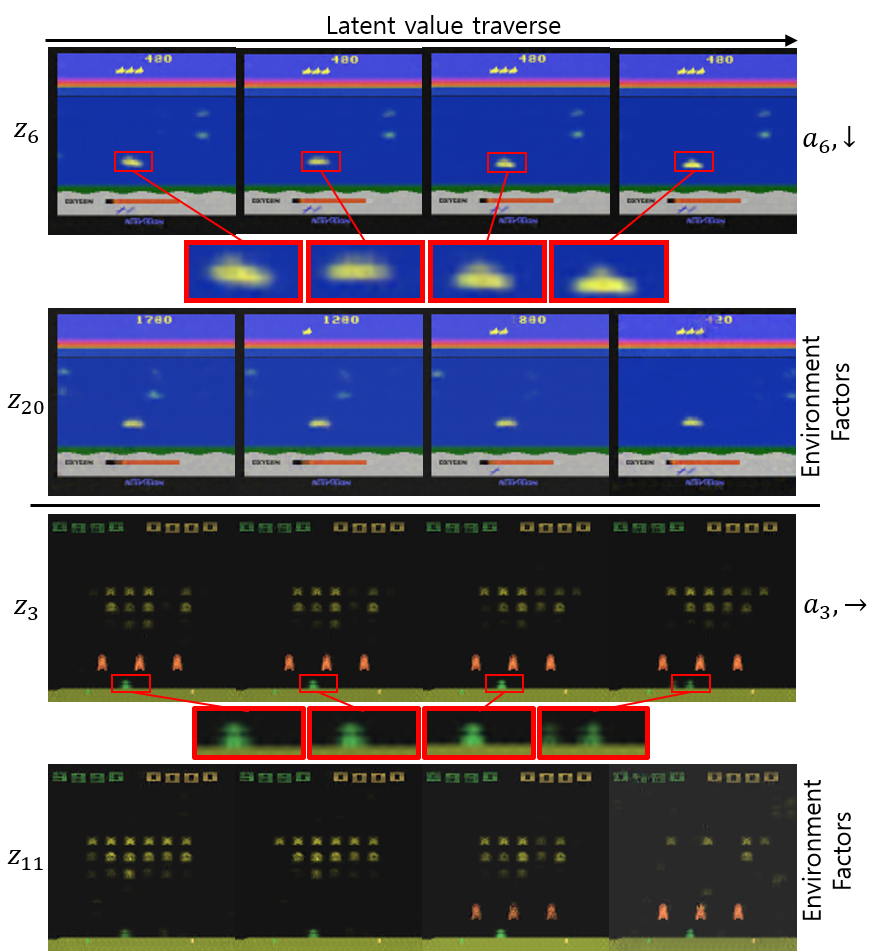}
\caption{The images are \nj{the estimated next states obtained by} traversing the latent vector $z \in \mathbb{R}^{20}$ learned by A2C policy and AC-$\beta$-VAE with $\beta$=10 and $\alpha$=0.001 on Atari game environments \textsc{Seaquest} (top) and \textsc{Space-Invaders} (bottom) with action spaces of $\mathbb{R}^{18}$ and $\mathbb{R}^{6}$, respectively. \nj{Because of a small movement per action, we have enlarged the ego at a fixed location (red box). }}
\label{fig:atari_2}
\end{figure}

\begin{figure*}[ht]
\centering  
  \begin{subfigure}{\linewidth}
    \centering
    \begin{subfigure}{.495\textwidth}
    \centering
    \captionsetup{justification=centering}
    \includegraphics[width = 0.9\linewidth]{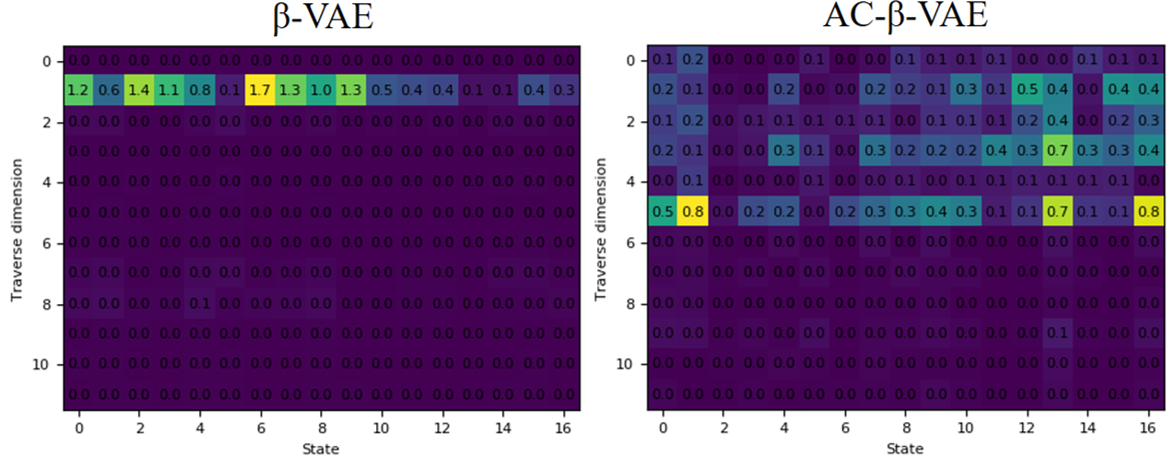}\\
    \vspace{-5pt}
    \caption{\textsc{Walker2d} ($a \in \mathbb{R}^6$, $z \in \mathbb{R}^{12}$, $a_{1:6} \rightarrow z_{1:6}$)}
    \end{subfigure}\hfill
    \begin{subfigure}{.495\textwidth}
    \centering
    \captionsetup{justification=centering}
    \includegraphics[width = 0.9\linewidth]{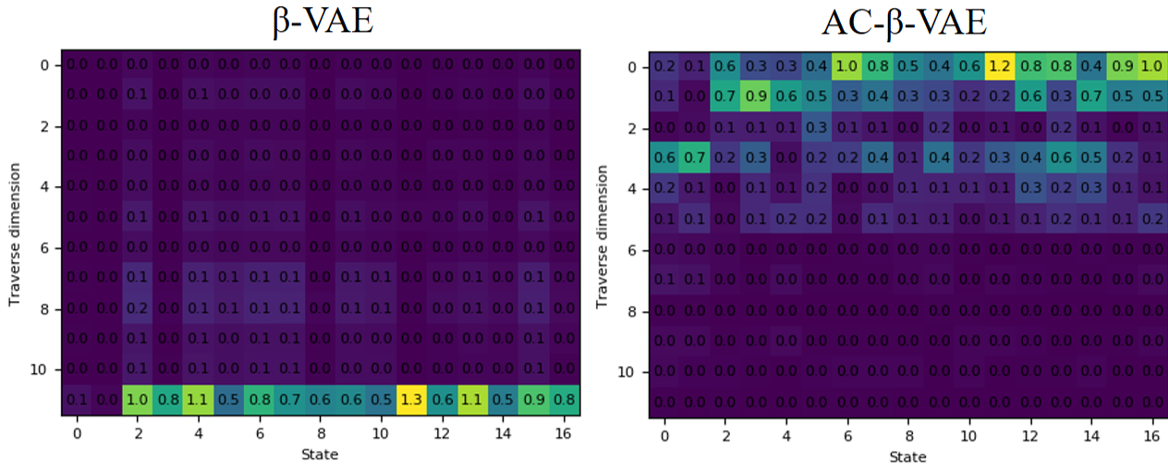}\\
    \vspace{-5pt}
    \caption{\textsc{Hopper} ($a \in \mathbb{R}^6$, $z \in \mathbb{R}^{12}$, $a_{1:6} \rightarrow z_{1:6}$)}
    \end{subfigure}\hfill
    \vspace{7pt}
  \end{subfigure}
  
  \begin{subfigure}{\linewidth}
    \centering
    \begin{subfigure}{.495\textwidth}
    \centering
    \captionsetup{justification=centering}
    \includegraphics[width = 0.9\linewidth]{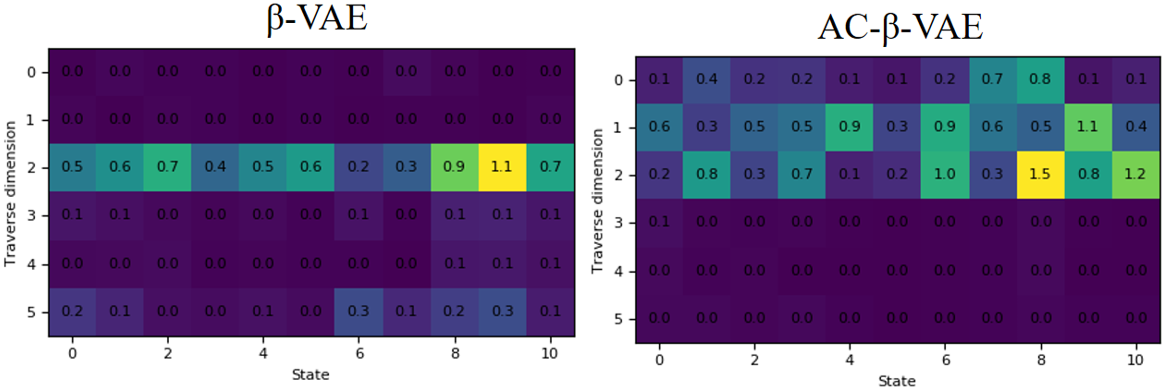}\\
    \vspace{-5pt}
    \caption{\textsc{Half-Cheetah} ($a \in \mathbb{R}^3$, $z \in \mathbb{R}^{6}$, $a_{1:3} \rightarrow z_{1:3}$)}
    \end{subfigure}\hfill
    \begin{subfigure}{.495\textwidth}
    \centering
    \captionsetup{justification=centering}
    \includegraphics[width = 0.9\linewidth]{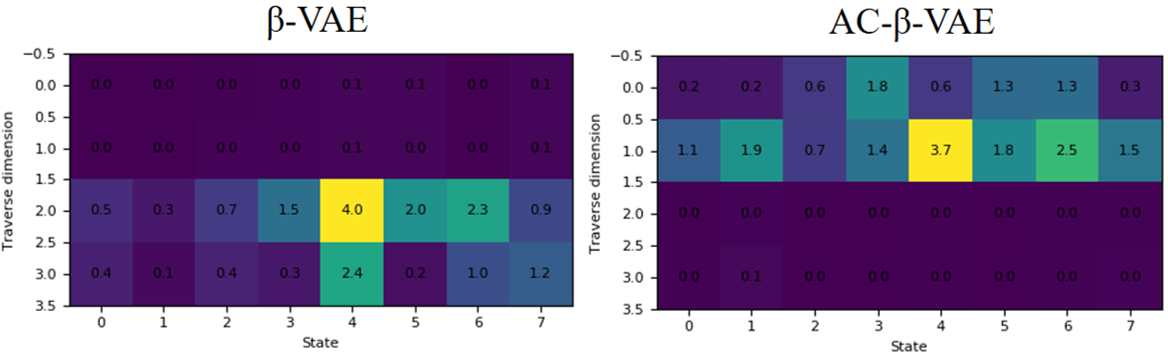}\\
    \vspace{-5pt}
    \caption{\textsc{Swimmer} ($a \in \mathbb{R}^2$, $z \in \mathbb{R}^4$, $a_{1:2}\rightarrow z_{1:2})$}
    \end{subfigure}\hfill
  \end{subfigure}
\caption{Traverse results in the MuJoCo environments. 
  The numbers in the boxes represent the standard deviations of each dimensional factor of the following state, $s_{t+1}$, when traversing the corresponding dimensional factor of the latent vector. Compared to the traverse for unmapped dimensions, the standard deviations of state values in the action-mapped dimensions are larger. Right arrows indicate action-mapping dimensional locations.}
\label{fig:trav_0}
\end{figure*}

\noindent \textbf{MuJoCo Environment} provides a physics engine system for rigid body simulations \cite{Todorov2012mujoco,Brockman2016openai}. 
Four robotics tasks  are engaged in our experiments: \textsc{Walker2d, Hopper, Half-Cheetah} and \textsc{Swimmer.}
A state vector represents the current status of a provided robotic figure, each factor of which is unknown of its physical meaning.

\nj{As an encoder and a decoder, we} have used a convolutional neural network (CNN) for Atari environments and fully-connected MLP networks for dSprites and MuJoCo environments.
For the stochastic policy network, we have used a fully-connected MLP. 
PPO and A2C are applied to optimize agent's policy for continuous control and discrete actions, respectively.
Most of hyper-parameters for the policy optimization are referred from the works of \cite{Schulman2017PPO,wu2017a2c}.

\subsection{Disentanglement \& Interpretability}
To demonstrate the disentanglement performance and interpretability of the proposed algorithm, 
we have experimented our method with $(s_t, a_t, s_{t+1})$ tuples from environments mentioned above.

Figure \ref{fig:traverse_de} and Table \ref{tab:quant_scores} illustrate the results for the dSprites environment.
The metric framework suggested in \cite{eastwood2018framework} with a random forest regressors are applied to present the quantitative results \nj{of disentanglement and completeness}. 
The tree depths are determined for the lowest prediction error of the validation set.
Since the metric system is based on the disentanglement for the conventional VAE and $\beta$-VAE, 
our metric results may not be strictly comparable to the ones reported in the original work.
In \nj{Fig. \ref{fig:traverse_de}(a) and (b)}, the VAE and $\beta$-VAE seem to struggle from learning the pattern between input $s_t$ and the output $s_{t+1}$ without any action constraint because of the randomness of the environmental squared object, creating relatively blurred reconstructions. 
Such excessive generalization in reconstructions \nj{results} in low scores in both disentanglement and completeness which means relatively low representational power to reproduce the ground truth variant factors.
Although the action conditions and the low-weighted $D_{KL}$ term allow \nj{AC-VAE} ($\beta$=1) reconstruct sharper images, its relatively low disentanglement pressure results in lower metric scores compared to AC-$\beta$-VAE \nj{($\beta$=20)}.

The results for the Atari environments in Figure \ref{fig:ego_worldmodel} and  Figure \ref{fig:atari_2} show that the latent vector trained with our method models the given environment \nj{successfully}. 
All the visited state space and learned behaviors can be projected by traversing \nj{each dimension of the latent vector}.
In that sense, our method can be considered as an action-conditional generative model.
Because AC-$\beta$-VAE \nj{can model the world} in an egocentric perspective, 
all \nj{the sequences} of \nj{(state-action-next state)} can be re-simulated.
Such trait may advance many RL methods since similar models are used for an exploration guidance \cite{tang2017exploration} or as the imagery rehearsals for training \cite{worldmodels}.



Figure \ref{fig:trav_0} shows the quantitative results of the traverse experiment on the MuJoCo environment.
Numbers on the heat-map represent the standard deviations for each dimension's state values when traversing dimensional factor.
The higher standard deviation value in the traverse of a specific dimension means the more effects the traversing dimension have on immediate state changes.
Unlike other environments, 
the MuJoCo environment has no environmental factors, 
and the current state is represented by the preceding movement of the given robotic body.
As shown in Figure \ref{fig:trav_0}, since the standard deviation of the state values during the traverse of 
the dimensions that are mapped with actions is larger than the unmapped ones, 
we can see the proposed algorithm is able to learn the disentangled action-dependent latent features.
However, it is limited from clear visual interpretation compared to the experimental cases in other environments 
because \nj{the} actions in the MuJoCo environment is defined as a continuous control of torques for all joints \nj{and} 
it is conjectured that the movement of one joint affects the whole status of the body.



\subsection{Controlling and Governing efficacy}
To verify the controllability of an agent's optimized efficacy,
we traverse the latent factors over the environment-specific range during an episode on the learned network.
In order to examine $s_{t+1}$, the environment output, 
the traversal is conducted before reparameterization ($\mu$ vector).
Furthermore, to get a clear view on the effect of action-mapped dimensions of the latent vector, 
we set all of the value of action mapped dimensions to zero except for the traversing one and those unmapped dimension of the latent vector.
These experiments are conducted on the Mujoco environments, and traverse range is set as [-5, 5] for every \nj{tasks}.

\begin{figure}[ht]
  \centering
  \includegraphics[trim={0.32cm 0.4cm 0.1cm 0.4cm},clip,width = 0.88\linewidth]{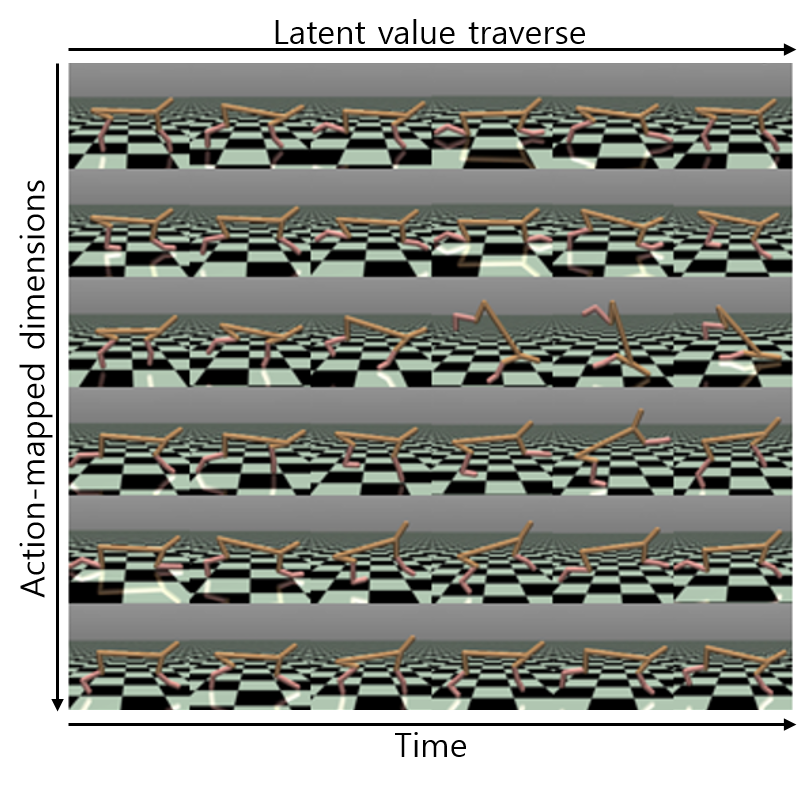}\\
   \caption{For \textsc{Half-Cheetah} environment with continuous control, latent behavioral factors can be interpreted by traversing latent values in time. 
   As a result, each action-mapped latent feature is responsible for a behavioral factor. 
}
  \label{fig:trav_render}
\end{figure}

The learned behavior in each latent dimension is also depicted in Figure \ref{fig:trav_render}. 
The resultant traverses of action-mapped dimensions on latent factors 
yield in behavioral movements that are combinations of multiple joint torque values.
Unlike in Atari environments with discrete action spaces, 
AC-$\beta$-VAE is constrained with various combinations of continuous action values during training simulations. 
When the policy network is optimized to accomplish a goal behavior such as walking, the action-mapped latent factors are learned to represent required behavioral components of spreading or gathering the legs. 
Therefore, 
$\mu$ vector represents variations in combinations of multiple joint movements, which allows for ease of visual comprehension on agent's optimized efficacy.
This clearly shows the possibility of governance over an RL agent's efficacy with human-level interpretations through controlling \nj{the values of the $\mu$ vector} in the latent space.

\begin{figure}[ht]
  \centering
  \includegraphics[width = 0.9\linewidth]{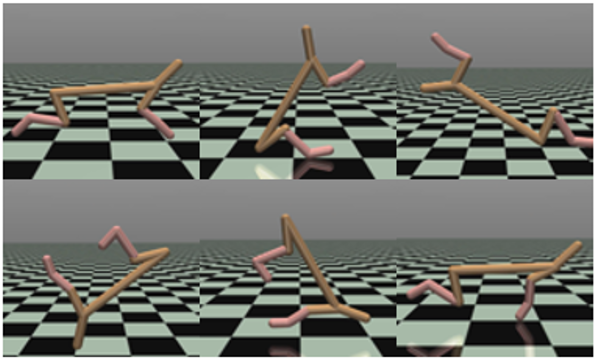}\\
  \caption{Example of governing the agent movement in MuJoCo environment of \textsc{Half-Cheetah}. 
  The robotic body is conducting a back-flip movement which is induced by controlling latent values at first and \gj{second} dimensions of 
  the learned $\mu$ vector \nj{shown in} Figure \ref{fig:trav_render} .}
  \label{fig:govern}
\end{figure}

We have taken the advantage of our transparent policy network and derived another behavior by controlling learned behavioral components.
An RL agent is able to learn with a reward function defined by human preference to perform, for example, a back-flip motion in \textsc{Hopper} environment \cite{humanpreference}. 
Showing a promising result of human enforcements on an RL model, 
our method enables governance over the agent's optimized behavior in \textsc{Half-Cheetah} environment.
After identification of behavioral components by traversing each \nj{element of the} $\mu$ vector, 
we are able to express another behavior of the agent, a back-flip in this case, as shown in Figure \ref{fig:govern}.



\section{Conclusion}
In this paper, we propose the action-conditional $\beta$-VAE (AC-$\beta$-VAE) which, for a given input state $s_t$ at time $t$, predicts 
next state $s_{t+1}$ conditioned \nj{on} an action $a_t$, sharing a backbone structure with a policy network during a deep reinforcement learning process.  
Our proposed model not only learns disentangled representations but distinguishes action-mapped factors and uncontrollable factors by partially mapping control-dependent variant features into the latent vector.
Since the policy network combined with the preceding encoder can be considered as one bigger policy network that takes raw states as inputs, with AC-$\beta$-VAE, we are able to build a transparent RL agent of which latent features are interpretable to human, overcoming 
conventional blackbox issue of Deep RL.
Such transparency allows human governance over the agent's optimized behavior with adjustments of learned latent factors.
We plan on the relevant studies for applications of the action-mapped latent vector.





\small
\bibliography{bib}

\end{document}